\documentclass[11pt,a4paper]{article}
\usepackage[hyperref]{acl2020}

\usepackage{url}

\usepackage{graphicx}  

\usepackage{times}
\usepackage{latexsym}

\usepackage{amsmath}

\usepackage{booktabs}       
\usepackage{multirow}
\usepackage{amssymb}
\usepackage{enumitem}
\usepackage{siunitx}

\usepackage{CJKutf8}

\aclfinalcopy 
\def\aclpaperid{473} 

\title{Guiding Variational Response Generator to Exploit Persona}

\author{
Bowen Wu$^1$, Mengyuan Li$^2$\thanks{* Contribution during the internship at Tencent.}, Zongsheng Wang$^1$, Yifu Chen$^3$\footnotemark[1], Derek F. Wong$^4$, \\
\textbf{Qihang Feng}$^1$, \textbf{Junhong Huang}$^1$, \textbf{Baoxun Wang}$^1$ \\
$^1$Platform and Content Group, Tencent \\
$^2$Peking University, Beijing, China \\
$^3$University of Chinese Academy of Sciences\\
$^4$NLP$^2$CT Lab / Department of Computer and Information Science, University of Macau \\
{\tt \small{\{jasonbwwu,jasoawang,careyfeng,vincenthuang,asulewang\}@tencent.com}}\\
{\tt \small{limengyuan@pku.edu.cn, chenyifu17@mails.ucas.ac.cn, derekfw@umac.mo}}
}

\date{}

\begin{document}
\maketitle
\begin{abstract}

Leveraging persona information of users in Neural Response Generators (NRG) to perform personalized conversations has been considered as an attractive and important topic in the research of conversational agents over the past few years. Despite of the promising progress achieved by recent studies in this field, persona information tends to be incorporated into neural networks in the form of user embeddings, with the expectation that the persona can be involved via End-to-End learning. This paper proposes to adopt the personality-related characteristics of human conversations into variational response generators, by designing a specific conditional variational autoencoder based deep model with two new regularization terms employed to the loss function, so as to guide the optimization towards the direction of generating both persona-aware and relevant responses. Besides, to reasonably evaluate the performances of various persona modeling approaches, this paper further presents three direct persona-oriented metrics from different perspectives. The experimental results have shown that our proposed methodology can notably improve the performance of persona-aware response generation, and the metrics are reasonable to evaluate the results.

\end{abstract}

\begin{figure*}
    \centering
    \includegraphics[width=1.\linewidth]{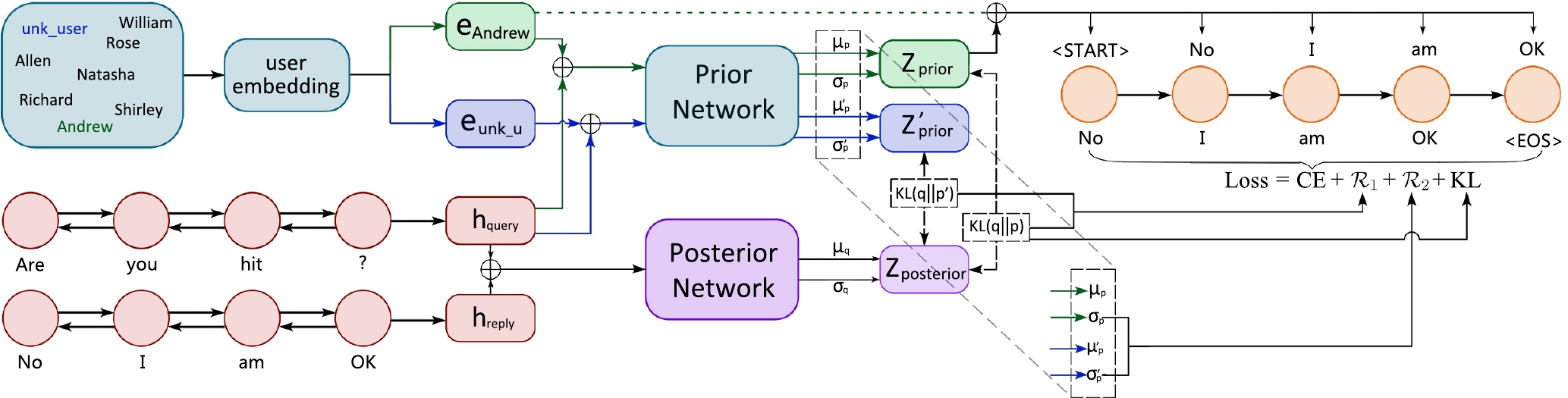}
    \caption{The architecture of the Persona-Aware Variational Response Generator (PAGenerator) described in this paper. $\oplus$ represents the concatenation of inputs and CE denotes the cross-entropy of predictions. The dotted arrow line indicates the connection is optional, and the default model named PAGenerator decodes with the user embedding.}
    \label{fig:fig_arch}
\end{figure*}

\section{Introduction}
\label{sec:intro}

As an essential research topic in generative conversational agents (a.k.a., chat-bots),
Persona Modeling is of great importance for such deep neural network based intelligent interactive systems~\cite{P16-1094,kottur2017exploring,wang2017steering}.
Apparently, user-personality-dependent responses provided by a chat-bot are able to significantly improve the consistency of its conversations, 
meanwhile, it is possible for users to flexibly customize the persona of a chat-bot based on some existent dialogues. 
As for the studies on this topic, with no doubt,
incorporating persona factors into End-to-End generative models is an attractive topic with great challenges. 

The current studies mainly focus on adopting the explicit meta-data of user profiles~\cite{qian2018assigning,chu2018learning} or character descriptions~\cite{P18-1205,mazare2018training,DBLP:conf/ijcai/SongZCWL19} to generate persona-aware responses. 
However, on one hand, user profiles are usually highly privacy-related 
and thus it is difficult to obtain such information from users practically. 
On the other hand, little correlation can be explicitly observed between such meta-data profiles and persona characteristics of users. 
Especially, those character descriptions, 
tailor-made for the persona-aware response generation with the great cost of manual work,
are only a variant of user profile innately in terms of different natural language forms. 

One of the reasonable and practically executable methodologies for introducing persona factors into conversation models is to adopt the real-valued user representation as a medium~\cite{P16-1094,kottur2017exploring,liu2018content,al2016conversational}. 
In particular, such user representations can be derived from users' historical dialog utterances with rich linguistic and personality information involved. 
Taking persona representations as the guidance for generating customized responses becomes a widely accepted methodology 
due to the recent development of deep latent variable models~\cite{zhao2017learning,shen2017conditional,P18-1104}.
However, for current models, without the explicit learning objectives or constraints, the user representation is adopted in a passive way to reduce the model loss and KL divergence via end-to-end learning.
In this case, it is highly possible that the employed embeddings will not work as effectively as expected.



Consequently, it is necessary to employ explicit guidance to help variational response generators sense persona.
From observations upon persona-contained dialogs, 
there exist intuitive characteristics for directing the optimization of the persona-aware variational response generation. 
Obviously, for a given user, 
the appropriately modeled and leveraged persona information can help to generate hidden variables semantically relevant with corresponding responses.
Besides, since users may have their own linguistic style, the adoption of personal information in NRG aims to have direct influence on the degree of linguistic (e.g. lexical and syntactic) convergence for a specific user.

This paper aims at exploring the explicit guidance to help the variational response generator exploit persona information hidden in the non-structured contents produced by the users, 
by utilizing intuitive characteristics of personalized conversations for model training. 
The contributions of this paper can be summarized as follows:
\begin{itemize}
    \item A persona-aware variational response generator is proposed to exploit persona while modeling the conversations.
    \item Based on the model, two regularization terms are presented to guide the model in encoding user information into the latent variables and converging to user-specific responses.
    \item Three discriminative metrics are further introduced to evaluate the capabilities of persona-aware response generators. 
\end{itemize}


\section{Approach}
\label{section:approach}

Based on the current progress on the development of latent variable models, we propose a persona-aware variational response generator to automatically exploit persona from conversations, and utilize such personal information to model the future conversation.
Besides, given that personal information can be exploited as optimization guidance to better modeling persona, we further introduce two regularization terms to guide the model learning.
In the following section, we first describe the general structure of PAGenerator, and then explain the two additional regularization terms.

\subsection{Persona-Aware Variational Response Generator}
\label{sec:approach_arch}

Utilizing latent variables in response generation has become a widely accepted methodology in NRG due to their Bayesian essence.
It helps to deal with external knowledge efficiently, e.g. Persona. 
Therefore, our proposed model is built based on the generation model with latent variables.
The overall architecture of the single turn persona-aware variational response generator proposed in this paper is illustrated in Figure~\ref{fig:fig_arch}.

Let $q, r, u$ stand for the query, the reply and the corresponding user of $r$, respectively, 
and $e_u$ stands for the embedding of user $u$. 
A bidirectional LSTM is first employed to encode the query and reply into fixed size vectors $h_{q}$ and $h_{r}$.
After that, the prior network (parametrized by $\theta$) takes $u_e$, $h_q$ as inputs to generate the distribution $p_{\theta}(z|q, u)$ of latent variable $z$. 
Meanwhile, $h_q$, $h_r$ are fed into a posterior network (parameterized by $\phi$) to compute $q_{\phi}(z|q, r)$.
As we adopt the assumption that $z$ follows isotropic Gaussian distribution, 
$p_{\theta}(z|q, u)$ and $q_{\phi}(z|q, r)$ are also normally distributed, such that:
\begin{equation}
\small
\begin{aligned}
p_{\theta}(z|q, u) &\sim \mathcal{N}(\mu_p, \sigma_p^2\textbf{I}) \\
q_{\phi}(z|q, r) &\sim \mathcal{N}(\mu_q, \sigma_q^2\textbf{I})
\end{aligned}
\end{equation}
where the means and variances are computed as follows:
\begin{equation}
\small
\label{equ:prior_net}
    \left[ \begin{array} { c } { \mu_p } \\ { \log(\sigma_p^2) } \end{array} \right]
    = W_p \left[ \begin{array} { c } { q } \\ { u } \end{array} \right] + b_p
\end{equation}
\begin{equation}
\small
\label{equ:post_net}
    \left[ \begin{array} { c } { \mu_q } \\ { \log(\sigma_q^2) } \end{array} \right]
    = W_q \left[ \begin{array} { c } { q } \\ { r } \end{array} \right] + b_q
\end{equation}
where $W_p$, $W_q$, $b_p$ and $b_q$ are the trainable parameters.
A sample of $z$ using the reparametrization trick~\cite{kingma2013auto} is then fed into the decoder as a part of input at each time step.

In addition, the bag-of-word (BOW) loss~\cite{zhao2017learning} is employed to tackle the latent variable vanishing problem, 
and PAGenerator is trained to maximize the variational lower-bound~\cite{chung2015recurrent,serban2017hierarchical}:
\begin{equation}
\label{equ:loss_1}
\small
\begin{aligned}
\mathcal{L}(\theta,\phi; q,r,&u) = {\mathrm{E}}_{q_{\phi}(z|q, r)}[\log p_{\theta}(r|z, q, u)]\\
- & KL(q_{\phi}(z|q, r)\|p_{\theta}(z|q, u)) \\
+ & \mathrm{E}_{q_{\phi}(z|q, r)}[\log p(r_{bow}|z, q, u)]
\end{aligned}
\end{equation}

\subsection{User Information Enhancing Regularization}
\label{sec:approach_reg1}

Ideally, we expect that the introduction of user embedding is fully utilized during model training.
However, due to the KL vanishing problem, 
the training of PAGenerator suffers from the hazard that the rapid decrease of $\mathcal{L}$ in Equation~\ref{equ:loss_1} might be attributed to the strong fitting capability of the decoder on the training data, 
rather than the involvement of user embedding.
Thus, we introduce a regularization term to promote the usage of user's hidden information in latent variables.

At the beginning, as illustrated in Figure~\ref{fig:fig_arch}, 
a general $unk\_u$ is introduced to represent the case for user unspecified.
Subsequently, taking the default user embedding $e_{unk\_u}$ as input, 
we obtain the KL divergence as $KL(q_{\phi}(z|q, r)\|p_{\theta}(z|q, unk\_u))$ from the network.
In this case, once the real user $u$ is introduced, 
a regularization term $\mathcal{R}_1(\theta,\phi; q,r,u)$ can be constructed as follows:
\begin{equation}
\label{equ:loss_2}
\small
\begin{aligned}
\mathcal{R}_1(\theta,&\phi; q,r,u) = \max(-\gamma_1, \\
& KL(q_{\phi}(z|q, r)\|p_{\theta}(z|q, u))\\
& - KL(q_{\phi}(z|q, r)\|p_{\theta}(z|q, unk\_u)) ) \\
\end{aligned}
\end{equation}
where $\gamma_1 \in \mathbb{R}$, $\gamma_1 > 0$, and $p_{\theta}(z|q, unk\_u) \sim \mathcal{N}(\mu_p', \sigma_p'^2\textbf{I})$.

It should be noted that, according to the equation above, 
the two prior distributions are generated from the same network with partially different inputs ($u$ VS. $unk\_u$),
and the regularization constrains the prior distribution with specified user to be closer to the posterior distribution.
Thus, the optimization encourages the utilization of user information and correspondingly inhibits the generated results from ignoring the user information. 
Meanwhile, $\mathcal{R}_1$ in our proposed model also alleviates the KL vanishing problem.

\subsection{Variance Controlling Regularization}
\label{sec:approach_reg2}

The BOW loss forces the latent variables to predict the bag-of-words in the response.
Therefore, the semantic distribution of $z$ is required to be capable of representing the topics and wording of the target response.
Besides, for a given query, 
the possible replies from a specific user should be more convergent to each other than those from an unknown user, 
due to each user's unique preference on the topics and wording.
Correspondingly, under the assumption that the distribution of $z$ represents the user's language preference,
the specification of user information is expected to reduce the entropy of the isotropic Gaussian distribution of $z$, 
reflected by a lower standard deviation $\sigma_p$.
On this basis, we introduce another regularization term $\mathcal{R}_2(\theta,\phi; q,r,u)$ to control the variance:
\begin{equation}
\label{equ:loss_3}
\small
\begin{aligned}
\mathcal{R}_2(\theta,&\phi;q,r,u) = \max(-\gamma_2, \sigma_p^2 - \sigma_p'^2) \\
\end{aligned}
\end{equation}
where $\gamma_2 \in \mathbb{R}$ and $\gamma_2 > 0$.
$\mathcal{R}_2$ prefers those $z$ with decrease $\ge \gamma_2$ in standard deviation $\sigma_p$ after specifying users,
and such decrease indicates the latent variables are more semantically convergent.

On this basis, we update the new training objective of PAGenerator as follows:
\begin{equation}
\label{equ:loss}
\small
\begin{aligned}
\mathcal{L}'(\theta,&\phi; q,r,u) = \mathcal{L}(\theta,\phi; q,r,u) \\
- & \mathcal{R}_1(\theta,\phi; q,r,u) - \mathcal{R}_2(\theta,\phi; q,r,u)
\end{aligned}
\end{equation}
By employing the two regularization terms to constrain the model training, 
$\mathcal{L}'(\theta,\phi; q,r,u)$ now also pays attention to the utilization of user information and language preference.

\section{Specified Evaluation Metrics of Persona NRG}
\label{section:metric}

In the previous section, two regularization terms are proposed to guide the model in the persona exploration.
However, we still lack effective persona-focused metrics to quantify how well one model is on learning persona.
The currently applied metrics for persona-aware NRG evaluation, such as perplexity and BLEU, are used to evaluate the plain NRG models~\cite{P16-1094,kottur2017exploring}.
Apparently, such metrics are inadequate to evaluate the capacity of a response generator on capturing persona.

Innately, an effective persona-aware response generator should be able to successfully identify and generate responses for users according to their language styles. 
Besides, the generated responses from different users should be diversified to each other in wording.
Considering these properties, 
we propose the following metrics to measure the level of persona-aware in response generators.

\subsection{Language Style Detection}
\label{section:metric_1}

It is important for a persona-aware response generator to identify a user's response from other user-irrelevant ones, 
by detecting the user's language style in responses.
In this subsection, we propose User-Relative-Rank (\textit{uRank}) to measure such capability.
Given a query-response-user triple $\{q, r, u\}$, a pre-trained seq2seq model $S2S$ and a model $M$ to be evaluated, 
we first generate $n$ user-irrelevant responses $\{r'_{i} | i \in [1,n]\}$ from $S2S$ using beam search.
For a desired persona-aware model $M$, it is expected to assign the ground truth response $r$ with a higher probability than other user-irrelevant ones $\{r'_{i} | i \in [1,n]\}$.
Thus, taking $S2S$ as reference, we set \textit{uRank} to be 1 if $M$ scores $r$ a higher ranking position among $r'_{i}$ than $S2S$, specifically:
\begin{equation}
\small
\begin{aligned}
& rank_{M} = | \{ i | P_M(r'_{i}) > P_M(r) \} | \\
& rank_{S2S} = | \{ i | P_{S2S}(r'_{i}) > P_{S2S}(r) \} | \\
& \textit{uRank} = 
\begin{cases}
    1 & \text{if \ } rank_{M} < rank_{S2S} \\
    0 & \text{otherwise}
\end{cases}
\end{aligned}
\end{equation}
where $P_m(r)$ and $P_{s2s}(r)$ are the probabilities of $\{q, r, u\}$ given by $M$ and $s2s$ respectively, $|X|$ presents the cardinal number of a set $X$,
and the lower score of either $rank_{M}$ or $rank_{S2S}$ indicates a better ranking position.
Overall, for model $M$, its average \textit{uRank} for different queries denotes the rate of rank-promoted ground-truth replies.

\subsection{Language Style Imitation}
\label{section:metric_2}
Apart from perceiving users' language styles, 
an effective persona-aware model should also be able to imitate language styles by generating responses satisfying users' language behaviors.
User-Language-Perplexity (\textit{uPPL}) is proposed to measure this property.

Given a user $u_i$, to conduct such metric, a statistical language model $LM_i$ is first trained using the user's utterances.
After that, for a generated response $r'$, 
its corresponding \textit{uPPL} is defined as the perplexity of $r'$ given by $LM_i$.
\textit{uPPL} quantifies the power of a persona-aware model on generating responses similar to users' history 
utterances.

\subsection{Diversity between Users}
\label{section:metric_3}

Finally yet importantly, due to the introduction of user information, given a query, 
we expect that responses for different users from a persona-aware model should be also diversified.
Therefore, Users-Distinct (\textit{uDistinct}) is proposed in this paper to capture such property. 
Given a query $q_i$ and $m$ different users $\{u_j\ | j \in [1,m]\}$,
we generate different responses $\{r'_j\ | j \in [1,m]\}$ for each user using $M$.
On this basis, Distinct-1 and Distinct-2~\cite{li2016diversity} of the response set $\{r'_j\ | j \in [1,m]\}$ are utilized to measure the in-group diversity of responses generated by $M$ within users.
\citet{P16-1094} also compare models through the case studies from the similar perspective.

\section{Experiments Setups}
\label{section:experiment_setups}

\subsection{Datasets}
\label{section:exp_data}
To evaluate the performance of our proposed method, 
we implement experiments on a Chinese Social Networking Service (SNS) corpus and the Cornell Movie Dialogues corpus~\cite{Danescu-Niculescu-Mizil+Lee:11a}.
The Chinese SNS corpus is crawled from a Chinese social network service Douban,\footnote{https://www.douban.com/group} 
containing totally 1,022,592 single-turn dialogues from 12,857 users;
while the Cornell Movie Dialogues corpus consists of conversations from movie scrips.
By cleaning up the Cornell corpus with the opensource script,\footnote{https://github.com/suriyadeepan/datasets/} we obtain 109,952 single-turn dialogues from 9,035 movie characters.
The training/test ratios for the two corpora are around 200:1 and 50:1, respectively.
Besides, for the Douban corpus, the mean, maximum, minimum, and the standard deviation values of the number of utterances for each user are 80, 1190, 33, and 49, respectively. Meanwhile, these statistics values are 14, 237, 4, and 22, correspondingly.

There are two main differences between the two datasets:
1) The scenes of conversations are different.
The dialogues in Douban are crawled from an open domain social media.
By contrast, since the characters in Cornell movie corpus are assigned with fixed personas, 
the language styles and habits of users are more templatized.
Besides, the language style in Cornell is more oral-like, with many personal pronouns.
2) The average number of utterances for each user of the Douban corpus is around 10 times more than that of Cornell.

\subsection{Model Variations}
\label{section:exp_baseline}

\begin{description}[leftmargin=0cm]
\setlength{\itemsep}{4pt}
\setlength{\parsep}{0pt}
\setlength{\parskip}{0pt}
\item[S2SA] Vanilla sequence-to-sequence model with attention~\cite{sordoni2015neural}.
\item[fact\_bias] S2SA with fact bias for persona modeling~\cite{P18-2050}. 
fact\_bias is originally proposed in NMT, it models user information as an additional bias vector learned through a factored model in the softmax layer.
\item[Speaker Model] Framework proposed by~\citet{P16-1094}. This model is similar to S2SA + fact\_bias, except that the user information is added as a part of decoder input rather than bias in the softmax layer.
\item[VAE] Standard Variational AutoEncoder for response generation~\cite{serban2017hierarchical}. In our experiment, we replace the utterance with the query only and apply the auxiliary BOW loss~\cite{zhao2017learning} in training. 
\item[CVAE] Conditional Variational AutoEncoder with user information as prior knowledge for modeling persona~\cite{zhao2017learning}. Similar to VAE, bag-of-words loss is applied in CVAE. 
\end{description}

For a fair comparison, we use the same configuration for all models. The size of word embedding and user embedding are respectively set to 300 and 128. All the user embeddings, including that of the unknown user, are initialized randomly and trained during the optimizing.
We employ a bi-directional LSTM of hidden size = 256 for encoding, 
and a LSTM of hidden size = 512 for decoding. 
For latent models, the dimension of $z$ is set as 128.

All models are optimized using Adam~\cite{kingma2014adam} with learning rate = $2e-4$ and batch size = 128.
For latent models, we also use KL annealing~\cite{bowman2016generating} (400,000 batches for Douban corpus and 100,000 batches for Cornell Movie corpus) to achieve better performance.

\subsection{Automatic Evaluation Metrics}
\label{section:exp_metric}
To thoroughly evaluate our systems, both standard and persona-focused metrics are employed in our experiments.
For standard metrics, we adopt unigram BLEU (BLEU-1)~\cite{papineni2002bleu} and Word Embedding metrics~\cite{liu2016not} including Embedding Average (Average), Vector Extrema (Extrema) and Greedy Matching (Greedy) to evaluate the semantics of generated responses with regards to ground truths.
We use the pretrained word embeddings from~\cite{song2018directional} for the Douban corpus and embeddings from~\cite{pennington2014glove} for the Cornell movie corpus.

The three proposed metrics (\textit{uRank}, \textit{uPPL} and \textit{uDistinct}) are adopted to measure the performance of capturing persona. 
For \textit{uPPL}, we use a bi-gram language model for perplexity computation.
Since the effectiveness of \textit{uPPL} relies on the quality of constructed user language models,
we pretrain the SLM with the whole training data and afterwards finetune it using each user's utterances. 
Besides, we drop the users with utterances less than 100 in Douban and 30 in Cornell.
The value of \textit{uRank}, which depends on the rankings of predicted probabilities of responses, is not stable for latent models due to the randomness on sampling $z$. 
Therefore, \textit{uRank} for each latent model is computed by running 10 rounds, 
so that we obtain 10 ranking results and their corresponding \textit{uRank}.
Then we average the obtained 10 \textit{uRank} as the final \textit{uRank} for each latent enhanced model.
The later experimental results show that \textit{uRank} for any latent model varies slightly around $\SI{ \pm 0.005}{}$ for each round.

\subsection{The Human Evaluation Criterion}
\label{section:exp_human}
For further comparisons,
we also use the crowdsourcing labeling resources of our organization to manually evaluate the relevance and the persona of generated responses.
Since the degree of persona reflected in the response is even more difficult to be judged by humans, 
we simplify the annotation into a ``yes or no'' task,
that is, annotators are only asked to decide whether the response can reflect persona for the given user.
Before that, the annotators have to read all the utterances of each user to learn the persona for judging.
Moreover, in practice, we limit the number of each user's sample utterances to 100.
However, the judgment is inevitably much more subjective.
Thus, for each sample, we recruit 11 annotators to label and make the final determination by voting.
The evaluation of relevance is relatively easy.
For the evaluation of relevance,
each query-response pair is cross-evaluated by 3 annotators, 
following the labeling criterion used in~\cite{xing2017topic,wang2018prospective}.
The details of data sampling and labeling are given in the \textbf{Supplementary Material}.





\begin{table*}[t]
\small
  \centering
    \begin{tabular}{l c ccc cccc}
        \toprule
        \multirow{2}{*}{Methods} & \multirow{2}{*}{BLEU} & \multicolumn{3}{c}{Embedding} & \multicolumn{4}{c}{Persona Metrics}\\
        \cmidrule(lr){3-5} \cmidrule(lr){6-9}
        & & Average & Extreme & Greedy & \textit{uRank} & \textit{uPPL} & \textit{uDist-1} & \textit{uDist-2}\\\midrule
        S2SA~\cite{sordoni2015neural}               & 0.29 & 0.834 & 0.615 & 0.666 & 0     & 200.4 & 0.115 & 0.113 \\
        fact\_bias~\cite{P18-2050}         & 0.29 & 0.840 & 0.618 & 0.671 & 0.022 & 202.3 & 0.091 & 0.101 \\
        Speaker Model~\cite{liu2016not}      & \textbf{0.31} & 0.837 & 0.621 & \textbf{0.674} & 0.023 & 163.6 & 0.183 & 0.199 \\
        VAE~\cite{serban2017hierarchical}                & 0.30 & 0.830 & 0.609 & 0.659 & 0.017 & 225.9 & 0.367 & 0.467 \\
        CVAE~\cite{zhao2017learning}               & \textbf{0.31} & 0.836 & 0.616 & 0.668 & 0.039 & 174.5 & 0.377 & 0.486 \\
        PAGenerator        & \textbf{0.31} & \textbf{0.845} & \textbf{0.622} & 0.670 & \textbf{0.044} & \textbf{\textit{153.3}} & \textbf{0.406} & \textbf{0.524} \\
        \bottomrule
  \end{tabular}%
  \caption{Evaluation results on Douban corpus. \textit{uDist} is the abbreviation for \textit{uDistinct} in the table.\label{table:res_douban}}
\end{table*}

\begin{table}[t]
\small
  \centering
    \begin{tabular}{l cccc}
        \toprule
        \multirow{2}{*}{Methods} & \multicolumn{4}{c}{Human Evaluation}\\
        \cmidrule(lr){2-5}
         & 0 & 1 & 2 & Persona\\\midrule
        S2SA          & 60.0\% & 35.0\% & 5.0\% & 1.6\% \\
        fact\_bias    & 70.0\% & 26.7\% & 3.3\% & 7.8\% \\
        Speaker Model & 53.2\% & 41.6\% & 5.2\% & 9.6\% \\
        VAE           & 58.3\% & 35.0\% & 6.7\% & 3.8\% \\
        CVAE          & 55.0\% & 38.8\% & 7.2\% & 12.2\% \\
        PAGenerator   & \textbf{\textit{51.7}}\% & 38.3\% & \textbf{10.0}\% & \textbf{13.4}\% \\
        \bottomrule
  \end{tabular}%
  \caption{Human labeled results upon generated responses of models trained on the Douban corpus, with the beam width of 10. The Fleiss' kappa~\cite{fleiss1973equivalence} on all annotations is around 0.65, which can be considered as ``substantial agreement".\label{table:res_human}}
\end{table}

\section{Results \& Analysis}

\subsection{Results on the Douban Corpus}
\label{sec:exp_douban}

We first report the performance on the Douban corpus.
The results of automatic evaluating metrics are illustrated in Table~\ref{table:res_douban},
numbers in bold mean that the improvement on that metric is statistically
significant over other methods (\textbf{p-value $\le$ 0.01}).
It is observed that the BLEU-1 scores of various models are relatively low and close to each other.
We attribute this to the fact that the semantics of possible responses for one query is highly diversified in terms of speaking styles and topics, 
there might be the situation that only a small portion of words share among the responses except those of high-frequency words~\cite{mou2016sequence,liu2016not}.
However, user enhanced models achieve higher BLEU-1 scores due to their capability in considering the preference of a user.

Furthermore, by comparing the performances on embedding metrics, we find that all models obtain decent scores, but none of the models outperform the others significantly.
Such phenomenons can also be observed in previous studies~\cite{serban2017hierarchical,wang2019enhancing}, since all the models generate responses semantically similar to the ground truths.
Despite this, PAGenerator achieves the highest score on average, 
which suggests the responses generated by PAGenerator are more semantically relevant to the ground truths.

While all models perform more or less the same on standard metrics, 
their experimental results on persona metrics are quite different. 
All persona-aware NRG models outperform S2SA and VAE which contain no user information on the \textit{uRank}, 
while the two variational models with user information significantly exceed the rest models.
It shows that persona-aware response generators, especially those exploiting user embeddings to generate latent variables, are more sensitive on identifying users' language styles.
Among all models with user modeling, our proposed PAGenerator achieves the highest \textit{uRank}.

The advantage of introducing persona information into NRG is also reflected by \textit{uPPL}.
The replies given by the three models employing user embeddings are more consistent with the user's language style, 
which indicates that user embedding is useful in learning language style automatically in an End-to-End NRG model.
By contrast, since S2SA with fact\_bias focuses on learning user's bias based on only unigrams, it struggles from achieving a high \textit{uPPL} which scores from bi-gram perspective.
Moreover, comparing the performance of CVAE to Speaker Model, 
it appears that utilizing latent variables in standard method cannot further improve \textit{uPPL}.
By contrast, the two new regularizations proposed for persona modeling can help PAGenerator generating replies with more specific persona, 
the \textit{uPPL} of which is reduced by $21.2$ points compared to CVAE.

\begin{table*}[t]
\small
  \centering
    \begin{tabular}{l c ccc cccc}
        \toprule
        \multirow{2}{*}{Methods} & \multirow{2}{*}{BLEU} & \multicolumn{3}{c}{Embedding} & \multicolumn{4}{c}{Persona Metrics}\\
        \cmidrule(lr){3-5} \cmidrule(lr){6-9}
        & & Average & Extreme & Greedy & \textit{uRank} & \textit{uPPL} & \textit{uDist-1} & \textit{uDist-2}\\\midrule
        S2SA~\cite{sordoni2015neural}               & 0.32 & 0.787 & 0.503 & 0.679 & 0     & 44.8 & 0.115 & 0.079 \\
        fact\_bias~\cite{P18-2050}         & 0.30 & 0.785 & 0.501 & 0.676 & 0.044 & 39.3 & 0.127 & 0.095 \\
        Speaker Model~\cite{liu2016not}      & \textbf{0.33} & 0.796 & 0.510 & 0.681 & 0.056 & 41.7 & 0.228 & 0.225 \\
        VAE~\cite{serban2017hierarchical}                & 0.25 & 0.780 & 0.490 & 0.670 & 0.058 & 45.6 & 0.122 & 0.114 \\
        CVAE\cite{zhao2017learning}               & 0.28 & 0.800 & 0.502 & \textbf{0.689} & 0.085 & 37.0 & 0.223 & 0.251 \\
        PAGenerator        & \textbf{0.33} & \textbf{0.814} & \textbf{0.514} & 0.687 & \textbf{0.114} & \textbf{\textit{32.2}} & \textbf{0.251} & \textbf{0.304} \\
        \bottomrule
  \end{tabular}%
  \caption{Comparison of different approaches on the Cornell Movie Dialogues corpus.\label{table:res_cornell}}
\end{table*}

\begin{table}[t]
\small
  \centering
    \begin{tabular}{cccc}
        \toprule
        Methods & \textit{uRank} & \textit{uPPL} & \textit{uDist-1/2}\\\midrule
        PAGenerator         & 0.114 & 32.2 & \textbf{0.251}~/~0.304 \\
        w/o $\mathcal{R}_1$ & 0.117 & \textbf{\textit{29.6}} & 0.209~/~0.246\\
        w/o $\mathcal{R}_2$ & \textbf{0.118} & 37.2 & \textbf{0.251}~/~\textbf{0.319} \\
        w/o UE              & 0.063 & 43.5 & 0.149~/~0.139 \\
        \bottomrule
  \end{tabular}%
  \caption{Ablation tests of PAGenerator on Cornell Movie Dialogue Corpus. "w/o" denotes PAGenerator does not contain the specific component, for example, "w/o UE" means the decoder of PAGenerator does not utilize the user embedding as input. \label{table:res_compare}}
\end{table}

As mentioned in previous sections, \textit{uDistinct} measures the diversity of the generated responses between different users. 
In general, latent models achieve higher \textit{uDistinct} than non-latent ones as the randomness brought by the latent variables.
Within latent models, the adoption of user information in CVAE only slightly improves its \textit{uDistinct} compared to VAE without user specification.
It indicates that user embeddings are ineffectively utilized in CVAE, 
and this is the motivation for us to propose new methods for variational response generator.
The notable improvement in \textit{uDistinct} can verify their effectiveness in exploiting persona.
The cases can further demonstrate such improvements in \textbf{Supplementary Material}.

Besides, the comparison among baseline models is consistent with the experiments in previous studies~\cite{P16-1094,P18-1104}, 
which indicates the proposed metrics are apposite for evaluating the capability of NRG models on capturing persona.

\subsection{Human Evaluation}
\label{sec:exp_human}

To further evaluate the quality of generated responses from each model more subjectively,
we also implement human labeling.
As shown in Table~\ref{table:res_human}, adjusting unigram distributions for users by fact\_bias reduces the quality of generated responses.
By contrast, all other models produce more high-quality replies comparing with S2SA.
Moreover, responses from PAGenerator achieve the best human evaluation result,
which indicates that the improvement of persona capturing of PAGenerator does not reduce correlation.

Meanwhile, in the last column, the trend of evaluated results on persona almost consists to those evaluated by proposed automatic evaluation metrics.
The PAGenerator outperforms other models, and some particular parts of replies generated by persona-aware models can reflect the personality.
Besides, due to the randomness, some responses given by S2SA and VAE are also labeled as persona-aware.
However, fewer high-quality responses generated by S2SA compared to VAE, 
and thus, the proportion of S2SA is even lower.


\subsection{Results on the Cornell corpus}
\label{sec:exp_cornell}

As shown in Table~\ref{table:res_cornell}, the overall trend of the experimental results on Cornell corpus is consistent with that on Douban corpus.
The models that are aware of the specified user outperform others slightly on BLEU and Embedding metrics.
Regards to persona metrics, the experimental results on Cornell corpus shows two main differences:
a) The Speaker Model does not perform that well on user language style detection and generation,
mainly because the training data of each user is less than that in Douban corpus. It is hard to automatically model the informative user embedding via target oriented learning without guidance.
By contrast, utilizing the KL divergence as the guidance in CVAE effectively improves the experimental results.
b) Due to the individual characteristics of movie characters, the user-embedding-enhanced models generate more diverse responses for different users, specially PAGenerator.

\subsection{Human Evaluation Results on the Cornell Corpus}

As shown in Table~\ref{table:res_human_en}, on the English dataset, the comparison results are almost consistent with that in Section~\ref{sec:exp_human}.
According to the judgment of annotators, our proposed model outperforms the others from both relevance and persona perspective.
However, influenced by insufficient training conversations, the overall quality of generated responses for the Cornell queries is not as good as the ones given for the Douban corpus.
We attribute this to the difference in the corpus size and the word distribution, which is described in Section~\ref{section:exp_data}.
In detail, the quality of Cornell is influenced by insufficient training conversations. 
By contrast, the persona is reflected more obviously with the help of more templatized language styles and habits of Cornell.

\subsection{Ablation Study}
\label{sec:exp_ablation}


To get a better intuition about how our proposed method works,
we implement the ablation tests to analyze the contribution of each component of PAGenerator in persona exploitation.
As illustrated in Table~\ref{table:res_compare}, adding the user embeddings as a part of decoder inputs brings positive improvements on all the persona-focused metrics.
Without UE, the parameter size of PAGenerator reduces considerably, which is harmful to the model on fitting target data.
Besides, without direct constraints from the decoder, user embeddings mainly act on reducing KL divergence rather than providing more informative latent variables.
Besides, without UE, PAGenerator also significantly outperforms VAE in all metrics, which demonstrates that $\mathcal{R}_1$ and $\mathcal{R}_2$ are indeed useful for guiding the latent variables to model the semantics under the query and users. 



Comparing the ablation results of w/o $\mathcal{R}_1$ with w/o $\mathcal{R}_2$,
we can conclude that both regularizations promote \textit{uRank} values.
However, PAGenerator w/o $\mathcal{R}_2$ only achieves a mediocre result on \textit{uPPL}, while only utilizing $\mathcal{R}_2$ damages the model's ability in generating diverse responses for different users.
We attribute this divergence to the trade-off between a) shared movie-style language between users and b) different language preferences among actors in the movie scripts.
Since $\mathcal{R}_1$ promotes the divergence of $z$ between the specified and unspecified users,
removing $\mathcal{R}_1$ raises the difficulty for the model to generate diverse responses toward different users, reflected by the low \textit{uDistinct} of w/o $\mathcal{R}_1$.
However, promoting diversity will more or less sacrifice the model's learning on the common shared movie-style patterns, which is vital in evaluating the language cohesion.
Therefore, the performance of PAGenerator only with $\mathcal{R}_1$ on \textit{uPPL} is less-than-ideal.
In contrast, since $\mathcal{R}_2$ emphasizes those patterns often used by a given user, 
it encourages the distribution of user information to be more aggregate.
These differences explain the opposite results of w/o $\mathcal{R}_1$ and w/o $\mathcal{R}_2$.


\begin{table}[ht]
\small
  \centering
    \begin{tabular}{l cccc}
        \toprule
        \multirow{2}{*}{Methods} & \multicolumn{4}{c}{Human Evaluation}\\
        \cmidrule(lr){2-5}
         & 0 & 1 & 2 & Persona\\\midrule
        S2SA          & 70.6\% & 27.5\% & 1.9\% & 1.4\% \\
        fact\_bias    & 72.2\% & 26.0\% & 1.8\% & 14.9\% \\
        Speaker Model & 62.2\% & 35.6\% & 2.2\% & 16.9\% \\
        VAE           & 65.0\% & 31.6\% & 3.4\% & 1.1\% \\
        CVAE          & 61.7\% & 34.0\% & 4.3\% & 21.6\% \\
        PAGenerator   & \textbf{\textit{61.5}}\% & 33.8\% & \textbf{4.7}\% & \textbf{22.8}\% \\
        \bottomrule
  \end{tabular}%
  \caption{Human evaluation results on the Cornell Corpus.\label{table:res_human_en}}
\end{table}

In conclusion, the user embedding is an important constraint for the PAGenerator, and $\mathcal{R}_1$, $\mathcal{R}_2$ can be considered to deploy for different purposes.
Furthermore, utilizing all components of PAGenerator described in Figure~\ref{fig:fig_arch} guarantees a more balanced and relatively best performance in all three evaluated persona exploiting abilities. 

\section{Related Work}
\label{sec:related}

\subsection{Persona-based Neural Models}

Persona-based neural conversation models can be categorized into two major research directions. One is to directly train a model from conversational data by considering the persona information~\cite{P16-1094,kottur2017exploring,wang2017steering,madotto2019personalizing}, while the other approach makes use of the profiles or side-information of users to generate the aligned responses~\cite{chu2018learning,qian2018assigning,P18-1205,mazare2018training,DBLP:conf/ijcai/SongZCWL19}. 
The work described in this paper belongs to the first research direction.
\citet{P16-1094} and~\citet{kottur2017exploring} enrich the models by training persona vectors directly and incorporating them into the decoder.
\citet{wang2017steering} propose three strategies to learn the language style instead of introducing new models.

Apart from the development of the Persona-based NRG models, 
recent researches also attempt to incorporate persona into neural machine translators.
\citet{P18-2050} propose to learn speaker-specific parameters for the bias term in the output to promote user preferring unigrams,
and~\citet{wuebker2018compact} introduce offset tensors to perform fine-tuning for each user.

\subsection{Variational Response Generator}

The variational response generators have drawn much attention recently, 
due to the observation that it can be flexible to include the effect from conditions based on its Bayesian architecture~\cite{zhao2017learning,shen2017conditional} and naturally promote diversity by involving sampling in the generate stage~\cite{serban2017hierarchical,du2018variational,shen2018improving}.
\citet{zhao2017learning} and~\citet{shen2017conditional} introduce frameworks taking various conditions to influence the model learning.
Afterwards,~\citet{P18-1104} include the emoji into the variational NRG model to generate responses with particular emotions.
Actually, these models~\cite{zhao2017learning,shen2017conditional,P18-1104} can also be deployed to the persona-aware response generation scenario.
The main difference is that the speaker of the response is unpredictable based on the query.
Thus, we have introduced the architecture proposed by~\citet{zhao2017learning} and modified it to adapt to the persona-aware generation, for the meaningful comparison. 
Especially,~\citet{DBLP:conf/ijcai/SongZCWL19} have utilized persona information into the CVAE architecture, except they focus on modeling and copying users' explicit profiles.

\section{Conclusions}
\label{sec:conclusion}
In this paper, we proposed a variational neural network to model the conversation as well as the persona of users.  
On the basis of the network, two regularization terms are designed to guide the model in emphasizing the importance of the hidden user information.
In addition, to better reflect the persona characteristics of the response generation model, three metrics have been introduced to quantify the level of persona of the generated responses.
Experimental results show that our approach significantly outperforms other baseline models and the proposed metrics are effective in evaluating the capabilities of models on generating persona-aware responses.

\section*{Acknowledgments}
\label{sec:acknowledgments}
This work was supported in part by the National Natural Science Foundation of China (Grant No. 61672555), and the Science and Technology Development Fund, Macau SAR (Grant No. 0101/2019/A2).
We sincerely thank the anonymous reviewers for their thorough reviewing and valuable suggestions.
\bibliography{acl2019}
\bibliographystyle{acl_natbib}


\appendix

\section{Details of Human Evaluation}
\label{Append:Evaluation}

\subsection{Labeling Dataset Preparation}

For each model with a given query set, three generated responses for each query are randomly sampled from the results given by the beam search with a beam size of 10.
Then, a total of 3,000 query-response pairs are prepared for labeling.

\subsection{Labeling Criterion of Relevance}
The labeling criterion for judging the relevance between the response and the given query is described as follows:

\noindent \textbf{0}: the quality of response is poor, it is either irrelevant to the query, or grammatically incorrect.

\noindent \textbf{1}: although the response itself is acceptable as a reply, its content is not informative and dull. 

\noindent \textbf{2}: the response is not only relevant and grammatically correct, but also informative or interesting. 




\section{Case Studies}
\label{sec:appendix}

\begin{figure*}[!ht]
    \centering
    \includegraphics[width=0.9\linewidth]{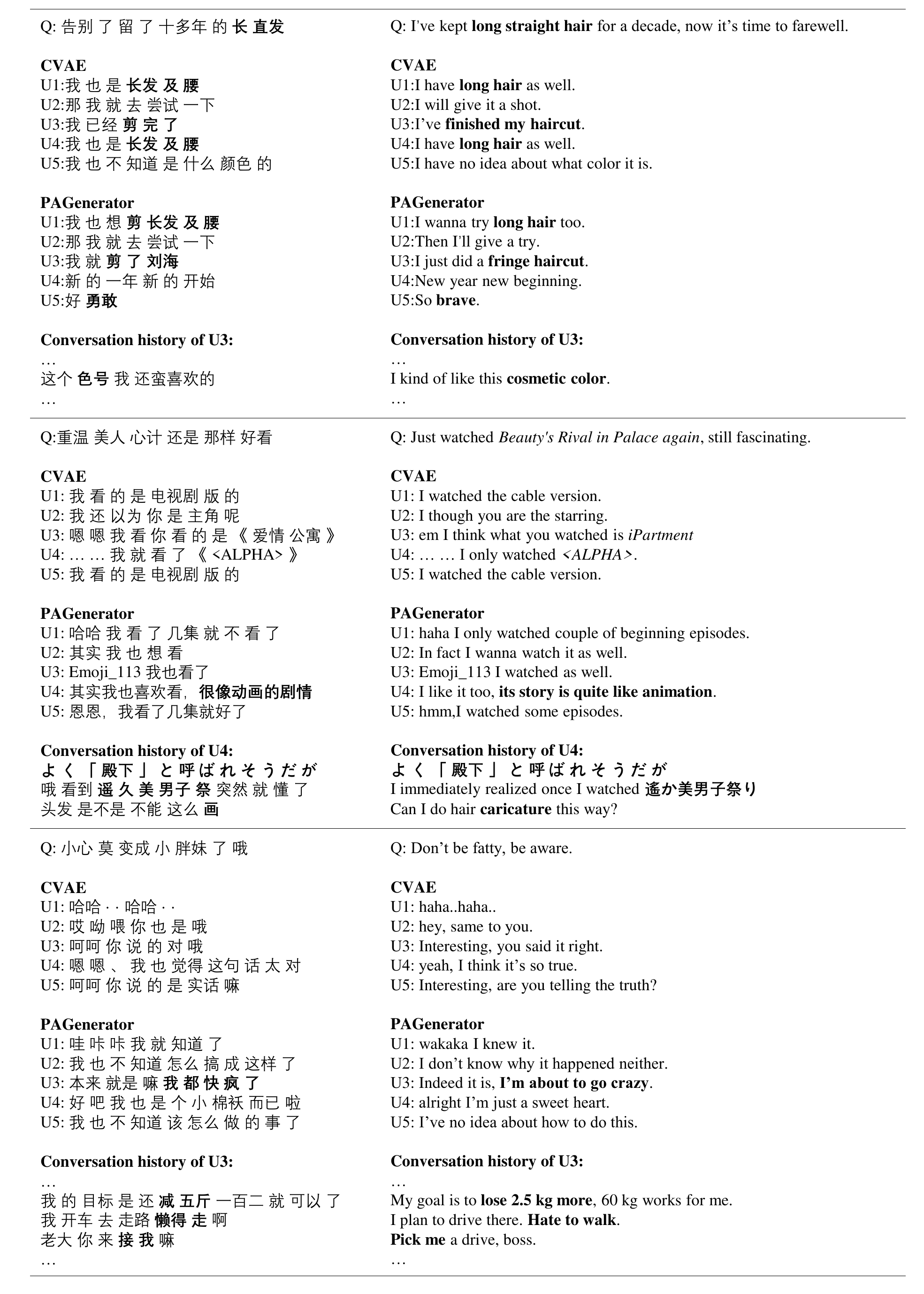}
    \caption{Comparisons of cases generated by CVAE and PAGenerator. Especially, we also give the utterances from conversation histories of some users (Conversation history of $U_k$, where $U_k$ denotes different users from different cases). The translated English version of the samples are listed on the right.}
    \label{fig:case1}
\end{figure*}

\begin{figure*}[!ht]
    \centering
    \includegraphics[width=0.9\linewidth]{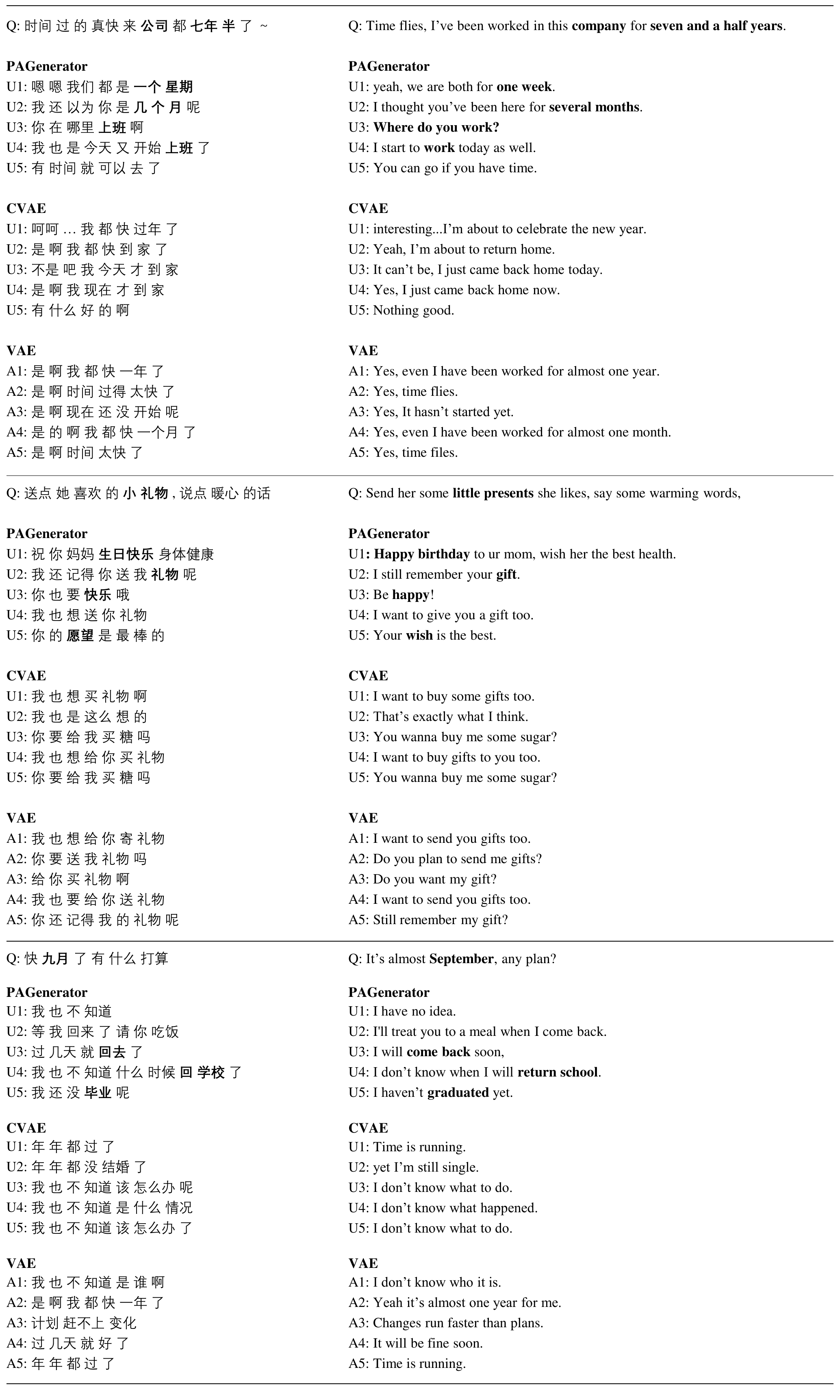}
    \caption{Cases for comparing the PAGenerator, CVAE and VAE. It should be noted that VAE have not adopted user information. The translated English version of the samples are listed on the right.}
    \label{fig:case2}
\end{figure*}

As shown in Figure~\ref{fig:case1}, we have selected three users whose utterances can reflect their implicit personal features. 
For example, the gender of user $U3$ in the first case is probably female. 
The user $U4$ in the second case is very possible to be an animation fun. 
According to the conversation history of user $U3$ in the last case, 
it can be inferred that the user is in the trouble of losing weight.
Correspondingly, from the responses generated by PAGenertor, 
we can observe that such implicit information are adopted by our proposed model to produce persona-aware results.

Figure~\ref{fig:case2} gives additional cases generated by 
PAGenertor, CVAE and VAE respectively oriented to the same given query.
Apparently, every independent user should have his/her own linguistic and personality characteristic. 
Thus, the results generated for different users are expected to maintain enough diversity.
According to the cases in Figure~\ref{fig:case2},
it can be seen that results of PAGenertor keep obvious diversity among different individuals, 
indicating its better capability of capturing persona of users.


\end{document}